\pdfoutput=1
\documentclass[11pt]{article}

\usepackage[]{ACL2023}

\usepackage{times}
\usepackage{latexsym}
\usepackage{algorithm,algorithmicx}% http://ctan.org/pkg/algorithm
\usepackage{algpseudocode}%http://ctan.org/pkg/algorithmicx
\usepackage{microtype}
\usepackage{booktabs}
\usepackage{multirow}
\usepackage{multicol}
\usepackage{array}
\usepackage{amsmath}
\usepackage{arydshln}
\usepackage{bbm}
\usepackage{graphicx}
\usepackage{graphics}
\usepackage{siunitx}
\usepackage{amsfonts}
\usepackage{makecell}
\usepackage{algorithm}

\usepackage{microtype}

\usepackage{inconsolata}

\usepackage{booktabs}
\usepackage{hyperref}
\usepackage{url}
\usepackage{graphicx}
\usepackage{xcolor}
\usepackage{amsmath}
\usepackage{amssymb}
\usepackage{color, colortbl}
\usepackage{makecell}
\usepackage{microtype}
\usepackage{xspace}

\usepackage[T1]{fontenc}
\usepackage[utf8]{inputenc}

\usepackage{microtype}

\usepackage{inconsolata}

\usepackage{times}
\usepackage{latexsym}
\usepackage{listings}
\usepackage[T1]{fontenc}
\usepackage[utf8]{inputenc}

\usepackage{microtype}

\usepackage{inconsolata}

\usepackage{booktabs}
\usepackage{hyperref}
\usepackage{url}
\usepackage{graphicx}
\usepackage{xcolor}
\usepackage{amsmath}
\usepackage{amssymb}
\usepackage{color, colortbl}
\usepackage{makecell}
\usepackage{microtype}
\usepackage{xspace}

\usepackage{array, tabularx, caption, boldline}
\usepackage{cellspace}
\usepackage{algpseudocode}  
\usepackage{amsmath}
\usepackage{multicol}  
\usepackage{multirow} 
\usepackage{flexisym}
\usepackage{graphicx}  %Required
\usepackage{array}
\usepackage{multicol}
\usepackage{multirow}
\usepackage{amsmath}
\usepackage{booktabs}
\usepackage{array}
\usepackage{soul}
\usepackage{vcell}
\usepackage{pbox}
\usepackage{enumitem}
\usepackage{graphics}
\usepackage{wrapfig}
\usepackage{makecell}

\usepackage{tabularx}
\makeatletter
\def\hlinewd#1{%
\noalign{\ifnum0=`}\fi\hrule \@height #1 %
\futurelet\reserved@a\@xhline}
\makeatother

\definecolor{darkgreen}{rgb}{0, 0.25, 0.}
\definecolor{darkblue}{rgb}{0, 0., 0.35}
\definecolor{LightCyan}{rgb}{0.88,1,1}

\title{Graph Reasoning for Question Answering with Triplet Retrieval}

\author{\makecell{Shiyang Li$^{1}$\thanks{~~Work was done during internship at Amazon.}, Yifan Gao$^{2}$, Haoming Jiang$^{2}$, Qingyu Yin$^{2}$, Zheng Li$^{2}$, Xifeng Yan$^{1}$ \\ Chao Zhang$^{3}$,  Bing Yin$^{2}$}\\
$^{1}$University of California, Santa Barbara \\ $^{2}$Amazon Inc. \\ $^{3}$Georgia Institute of Technology\\
\texttt{\{shiyangli,xyan\}@cs.ucsb.edu} \\
\texttt{\{yifangao,jhaoming, qingyy, amzzhe, alexbyin\}@amazon.com} \\
\texttt{chaozhang@gatech.edu}}

\begin{document}
\maketitle
\begin{abstract}
Answering complex questions often requires reasoning over knowledge graphs (KGs). State-of-the-art methods often utilize entities in questions to retrieve local subgraphs, which are then fed into KG encoder, e.g. graph neural networks (GNNs), to model their local structures and integrated into language models for question answering. However, this paradigm constrains retrieved knowledge in local subgraphs and discards more diverse triplets buried in KGs that are disconnected but useful for question answering. In this paper, we propose a simple yet effective method to first retrieve the most relevant triplets from KGs and then rerank them, which are then concatenated with questions to be fed into language models. Extensive results on both CommonsenseQA and OpenbookQA datasets show that our method can outperform state-of-the-art up to 4.6\% absolute accuracy.
\end{abstract}

\section{Introduction}

Answering complex questions is a challenging task since it often requires world knowledge and reasoning capability of underlying models \citep{Li2019TeachingPM,yasunaga-etal-2021-qa,zhang2022greaselm}. Pre-trained language models, e.g. BERT \citep{devlin-etal-2019-bert} and RoBERTa \citep{liu2019roberta}, have shown promising results by fine-tuning on downstream question answering tasks. However, world knowledge and reasoning of these models are learned from unstructured data, e.g. Wikipedia text, and are still limited \citep{Li2019TeachingPM,Petroni2019LanguageMA}.

On the other hand, there exist large-scale knowledge graphs (KGs), e.g. Freebase \citep{Bollacker2008FreebaseAC} and ConceptNet \citep{Speer2016ConceptNet5A}, capturing world knowledge explicitly by triples to record relations between entities \citep{zhang2022greaselm}. However, how to effectively integrate KGs into language models for question answering is still an open research problem. \citet{Li2019TeachingPM,Yu2020JAKETJP,Ye2019AlignMA,zhang-etal-2019-ernie,moiseev-etal-2022-skill} focus on utilizing KGs to construct distant supervision signals for continuous pre-training, however, KGs are often dynamic in practice and it is often hard to edit knowledge in models without further training, limiting their usage. \citet{bosselut-etal-2019-comet,wang-etal-2020-connecting} linearize reasoning paths in KGs and train language models on them to generate novel knowledge triplet during inference. However, KGs are discarded after training and language models can hallucinate false world knowledge \citep{Ji2022SurveyOH}. 

\citet{lin-etal-2019-kagnet,feng-etal-2020-scalable,yasunaga-etal-2021-qa,zhang2022greaselm,jiang-etal-2022-great,wang2022gnn} instead first recognize entities in questions and link them to KGs to retrieve subgraphs as additional input besides questions. However, this paradigm constrains retrieved knowledge in local subgraphs and discards more diverse triplets buried in KGs that are disconnected but useful for question answering. In addition, they require extra KG encoders with parameters trained from scratch besides standard language models, limiting model performance when training data is limited.

Recently, there have been growing interests to convert KG as a list of passages represented as natural languages. \citet{oguz-etal-2022-unik,Ma2021OpenDQ} convert KG triples into texts and combine these KG-converted texts with heterogeneous resources, e.g. tables and unstructured Wikipedia documents, as passages and achieved state-of-the-art performance for open-domain question answering. \citet{li-etal-2022-knowledge} follow this line of work and utilize KG and unstructured documents for knowledge-grounded dialogue generation. \citet{Zha2021InductiveRP} linearize reasoning paths in KGs for relation prediction and achieve state-of-the-art performance.

\begin{figure*}[t!]
  \centering
  \includegraphics[scale=0.50]{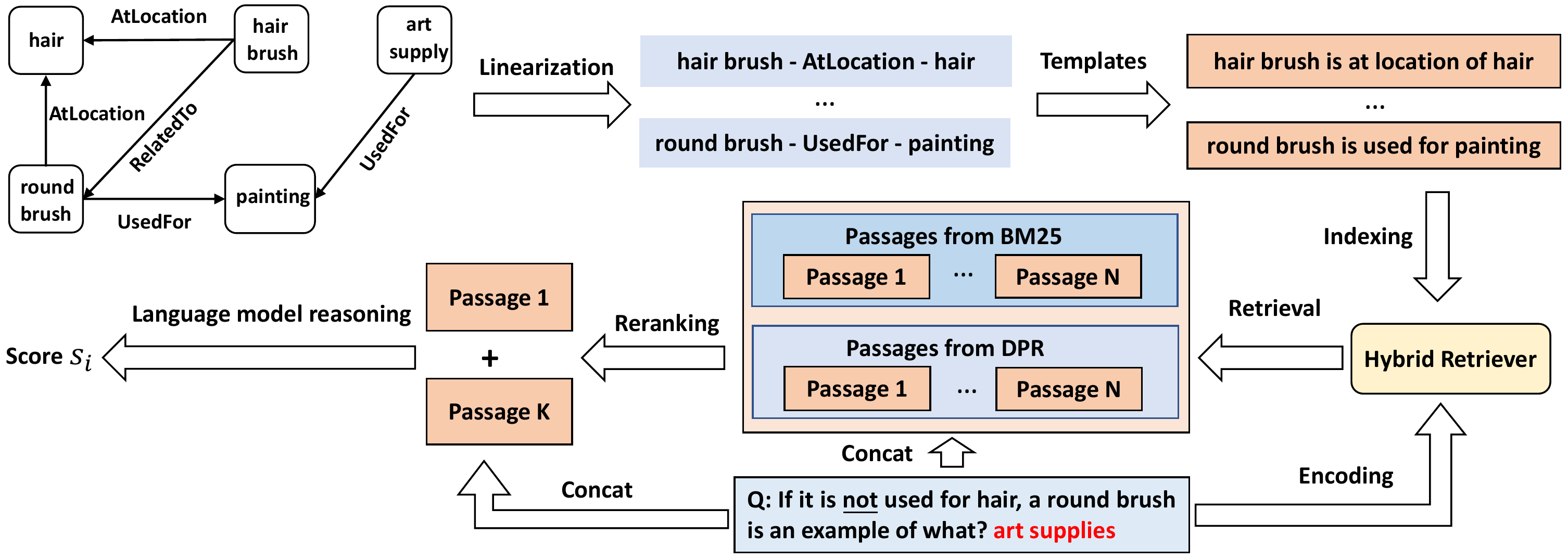}
  \caption{Overview of our framework. The exemplar KG is from \citet{yasunaga-etal-2021-qa}.}
  \label{fig:pipeline}
\end{figure*}

In this paper, we propose to conduct reasoning over KGs for question answering with triplet retrieval following \citet{oguz-etal-2022-unik,Ma2021OpenDQ,li-etal-2022-knowledge,Zha2021InductiveRP}. The overall pipeline of proposed method is shown in Figure \ref{fig:pipeline}. Specifically, we first linearize KG into triplets and convert them into passages by templates and directly retrieve the most relevant ones by questions with both sparse BM25 \citep{Robertson1994OkapiAT} and Dense Passage Retriever (DPR) \citep{karpukhin-etal-2020-dense}. We then rerank these passages by pre-trained cross-encoders \citep{reimers-gurevych-2019-sentence}. Finally, most relevant passages and questions are linearly concatenated and fed into pre-trained language model for question answering. This paradigm has several advantages compared to recent state-of-the-art \citep{yasunaga-etal-2021-qa,zhang2022greaselm,jiang-etal-2022-great,wang2022gnn}: (1) it is simple yet effective and can outperform state-of-the-art complicated question answering systems up to 4.6\% absolute accuracy (2) it does not need extra KG encoders trained from scratch, e.g. GNNs \citep{gnns2009,veličković2018graph}, and simply fuses knowledge passages and questions for question answering by standard language models. 

\section{Methodology}

\subsection{Problem setup}
We focus on multi-choice question answering (MCQA) tasks requiring model reasoning capability. Specifically, for each instance in a MCQA dataset, we have a question $q$ and a candidate choice set $C=\{c_{1}, c_{2}, \cdots, c_{n}$\}. We also assume that we have access to a knowledge graph $G$, which provides possibly relevant knowledge to answer each question. Given an example $(q,C)$ and a knowledge graph $G$, we aim to find the correct answer $c^{\star} \in C$.

\subsection{Knowledge graphs as passage corpus} \label{sec:convert2passage}

Knowledge graph $G$ can be represented as a list of triplets $P$. For each triplet $p \in P$, we convert it into natural language passage $d$ by templates so that relevant knowledge can be better retrieved. Specifically, for each triplet $p \in P$, it has a head entity $h$, a relation $r$ and a tail entity $t$. We map $r$ into $r_{p}$ and $d$ is formed by linearly concatenating <$h, r_{p}, t$>. For example, if we have a triplet <"hair brush", "AtLocation", "hair">, we map "AtLocation" into "is at location of" and form passage $d$ as "hair brush is at location of hair". Consequently, knowledge graph $G$ is converted into passage corpus $D$.

\subsection{Hybrid passage retrieval}
We retrieve passages from corpus $D$ for an MCQA example $(q,C)$ with hybrid (i.e., both sparse and dense) retrievers since they are complementary \citep{karpukhin-etal-2020-dense,Ma2021ARS}. For sparse retriever, we utilize BM25 to index $D$ and $(q,c_{i})$ to retrieve $N$ passages from it. For dense retriever, we use DPR due to its strong performance in open domain question answering \citep{karpukhin-etal-2020-dense}. DPR embeds queries with question encoder and passages with passage encoder into low dimensional dense vectors, and retrieval can be efficiently done through FAISS library \citep{johnson2017faiss} on GPUs. Similar to sparse retriever, we utilize $(q,c_{i})$ to retrieve $N$ passages from corpus $D$ for DPR. The total number of passages returned by BM25 and DPR is $2N$.

\subsection{Reranking}

We further rerank $2N$ passages retrieved by hybrid retriever with pre-trained cross-encoder \citep{reimers-gurevych-2019-sentence}. Specifically, for each passage $p$ among retrieved $2N$ passages, we concatenate query $(q,c_{i})$ and passage $p$ as the input of pre-trained cross-encoder. For each input, it will output a scalar value between 0 to 1. These scalar values are then used as reranking scores for $2N$ passages.

\subsection{Language model reasoning} \label{sec:lmr}

After reranking, we choose top $K$ passages $P_{K}$ and concatenate them along with question $q$ and choice $c_{i}$, which we cast as input of pre-trained language model (PLM). For input <$q, c_{i}, P_{K}$>, PLM will output contextual representation vector $\mathbf{h_{i}}$, which is then fed into a multi-layer perceptron (MLP) to output a scalar value $s_{i}$,
\begin{align}
\label{equation:1}
\mathbf{h_{i}} &=\text{PLM}(q,c_i,P_{K}),  \\
s_{i} &= \text{MLP}(\mathbf{h_{i}}),
\label{equation:2}
\end{align}
where $s_{i}$ is the prediction score of choice $c_{i}$ to be correct. During training, we calculate score $s_{i}$ for each choice $c_{i} \in C$ and normalize them with softmax function. After that, models are trained to maximize scores of correct choices with standard cross-entropy loss between predictions and ground truth labels. During inference, we calculate score $s_{i}$ for each choice $c_{i} \in C$ and select the one with the highest score as the predicted answer of question $q$.

\section{Experiments}

\subsection{Experimental setups}

We evaluate our method on two question answering datasets requiring model reasoning capability.

(1) \textbf{CommonsenseQA} \citep{talmor-etal-2019-commonsenseqa} is a 5-way multi-choice question answering dataset that requires common sense reasoning. Since its test set is not publicly available, we report in-house split  \citep{lin-etal-2019-kagnet} for comparisons with baselines.

(2) \textbf{OpenbookQA} is a 4-way multi-choice question answering dataset requiring multi-hop reasoning on scientific knowledge \citep{mihaylov-etal-2018-suit}. It has 4957/500/500 questions for training/development/test set split, respectively, and we report results on its test set.

\paragraph{Knowledge graph.} For knowledge graph, we utilize ConceptNet \citep{Speer2016ConceptNet5A}, which is a multi-relational and multi-lingual general knowledge graph storing world common sense knowledge. We first extract English triplets, clean them following \citep{yasunaga-etal-2021-qa} and convert them into natural language sentences as described in section \ref{sec:convert2passage}, resulting in 2,180,391 passages after data prepossessing. We defer details of relation mapping into Appendix \ref{sec:relation_mapping}.

\paragraph{Retrievers.} 
For sparse retriever, we utilize implementation of BM25 from \texttt{rank-bm25} python package \footnote{\url{https://github.com/dorianbrown/rank_bm25}} with default hyperparameters. For dense retriever, we utilize official pre-trained checkpoint \footnote{\url{https://dl.fbaipublicfiles.com/dpr/checkpoint/retriver/multiset/hf_bert_base.cp}} from DPR github repository \footnote{\url{https://github.com/facebookresearch/DPR}}.

\begin{table*}[t]
\centering
\small
%\scalebox{0.9}{
\begin{tabular}{p{0.6\columnwidth}ccccccc}
    \toprule
        \multirow{2}{*}{\textbf{Methods}}&
        \multicolumn{2}{c}{CommonsenseQA} &  \multirow{2}{*}{OpenbookQA}  \\
        \cmidrule(lr){2-3} 
        &  IHdev-Acc &  IHTest-Acc & \\
        \midrule
        RoBERTa-large (w/o KG)  & 73.07 (±0.45)$^{\dag}$ & 68.69 (±0.56)$^{\dag}$ & 64.80 (±2.37)$^{\dag}$\\
        \midrule
        RGCN \citep{Schlichtkrull2017ModelingRD} & 72.69 (±0.19)$^{\dag}$ & 68.41 (±0.66)$^{\dag}$ &  62.45 (±1.57)$^{\dag}$\\
        GconAttn \citep{Wang2018ImprovingNL}  & 72.61( ±0.39)$^{\dag}$ & 68.59 (±0.96)$^{\dag}$ &  64.75 (±1.48)$^{\dag}$\\
        KagNet \citep{lin-etal-2019-kagnet}  & 73.47 (±0.22)$^{\dag}$ & 69.01 (±0.76)$^{\dag}$ & -\\
        RN \citep{Santoro2017ASN}  & 74.57 (±0.91)$^{\dag}$ & 69.08 (±0.21)$^{\dag}$ & 65.20 (±1.18)$^{\dag}$\\
        MHGRN \citep{feng-etal-2020-scalable}  & 74.45 (±0.10)$^{\dag}$ & 71.11 (±0.81)$^{\dag}$ & 66.85 (±1.19)$^{\dag}$\\
        QA-GNN \citep{yasunaga-etal-2021-qa}  & 76.54 (±0.21)$^{\dag}$ & 73.41 (±0.92)$^{\dag}$ & 67.80 (±2.75)$^{\dag}$\\
        GreaseLM \citep{zhang2022greaselm}  & 78.5 (±0.5)$\star$ & 74.2 (±0.4)$\star$ & 66.9$\mathparagraph$\\
        SAFE \citep{jiang-etal-2022-great}  & - & 74.03$\star$ & 69.20$\star$\\
        GSC \citep{wang2022gnn}  & \underline{79.11} (±0.22)$^{\dag}$ & \underline{74.48} (±0.41)$^{\dag}$ & \underline{70.33} (±0.81)$^{\dag}$\\
        \midrule
        \textbf{Ours} & \textbf{79.80} (± 0.25) & \textbf{74.97} (± 0.56) & \textbf{74.93} (± 0.90)\\  
    \bottomrule
\end{tabular}%}
\caption{Performance comparison in accuracy (\%) on both CommonsenseQA and OpenBookQA datasets. We report the average results over three random seeds along with standard deviation on IHdev and IHTest \citep{lin-etal-2019-kagnet} for CommonsenseQA dataset and test set performance on OpenbookQA dataset. Best results are bold and second best ones are underlined. $\dag$: results from \citet{wang2022gnn}. $\star$: results from their original papers. $\mathparagraph$: results from \citet{yasunaga2022dragon}.}
\label{tab:full_results}
\end{table*}

\paragraph{Reranking.} 

We rerank retrieved passages from BM25 and DPR using pre-trained cross-encoder checkpoint from \texttt{sentence-transformers} package \footnote{\url{https://www.sbert.net/docs/pretrained_cross-encoders.html}}. Specifically, for CommonsenseQA dataset, we use pre-trained checkpoint \texttt{cross-encoder/ms-marco-MiniLM-L-12-v2} on MS MARCO dataset for passage ranking \footnote{\url{https://github.com/microsoft/MSMARCO-Passage-Ranking}} while for OpenbookQA dataset, we use pre-trained checkpoint \texttt{cross-encoder/stsb-roberta-large} on semantic textual similarity task \citep{cer-etal-2017-semeval}.

\paragraph{Language model reasoning.} 
Following \citet{yasunaga-etal-2021-qa,zhang2022greaselm,jiang-etal-2022-great,wang2022gnn}, we utilize RoBERTa-large \citep{liu2019roberta} to reason over passages and questions although our framework is model-agnostic. Specifically, question $q$, choice $c_{i}$ and passage list $P_{K}$ are linearly concatenated with special tokens among them and fed into models detailed in section \ref{sec:lmr} to predict choice score. 

We defer more implementation and training details of our method into Appendix \ref{sec:appendix}.

\paragraph{Baselines.}  
We experiment to compare our method to various baselines with extra KG-encoders, including RN \citep{Santoro2017ASN}, GconAttn \citep{Wang2018ImprovingNL}, RGCN \citep{Schlichtkrull2017ModelingRD}, KagNet \citep{lin-etal-2019-kagnet}, MHGRN \citep{feng-etal-2020-scalable}, QA-GNN \citep{yasunaga-etal-2021-qa}, GreaseLM \citep{zhang2022greaselm}, SAFE \citep{jiang-etal-2022-great} and GSC \citep{wang2022gnn}. For these baselines, RoBERTa-large \citep{liu2019roberta} is used for both CommonsenseQA and OpenbookQA. We also include RoBERTa-large fine-tuning only baseline without access to extra KG to show the effectiveness of our method. For all experiments in this work, we utilize accuracy (\%) as our evaluation metric.

\subsection{Main results}

As shown in Table \ref{tab:full_results}, our method can consistently outperform state-of-the-art on both CommonsenseQA and OpenbookQA datasets. For CommonsenseQA, our method' test performance can outperform fine-tuned RoBERa-large without KG 6.28\% absolute accuracy and outperform best baseline GSC with KG 0.49\%. On the smaller dataset OpenbookQA, our method' improvement is larger and can outperform fine-tuned RoBERa-large without KG 10.13\% absolute accuracy and outperform best baseline GSC with KG 4.60\% accuracy. Note that a key difference between our method and RoBERa-large without KG baseline is that our model also takes additional retrieved passages as input without introducing any extra parameters but can outperform various state-of-the-art methods with extra KG encoders. These consistent results indicate that our simple method can integrate knowledge into language models effectively.

\subsection{Ablation study}
\begin{table}[tb]
% \hspace{-2mm}
\centering
\small
%\scalebox{0.7}{
\begin{tabular}{lcc}
\toprule
Method          & CommonsenseQA     & OpenbookQA  \\
\midrule
GSC (best baseline) & 74.48(±0.41) & 70.33(±0.81) \\
\midrule
\textbf{Ours}   & \textbf{74.97} (± 0.56) & \textbf{74.93} (± 0.90) \\
~~- BM25  & 71.45 (±0.17) & \underline{72.47} (± 0.57) \\
~~- DPR  &  \underline{74.81 }(±1.35) & 71.73 (± 0.25)  \\
~~- Reranking  &  73.87 (±0.85)  & 70.93 (± 1.16) \\
\bottomrule
\end{tabular}
% \vspace{-2mm}
\caption{Results of removing BM25, DPR and Reranking module on both CommonsenseQA and OpenbookQA dataset.}
% \vspace{-2mm}
\label{tab:ablation}
\end{table}

We further ablate our method by removing BM25 retriever, DPR retriever and reranking module. Note that when we remove reranking module and we use the average score of BM25 and DPR if the same passage is retrieved; otherwise, following \citep{Ma2021ARS}, if a passage $p$ from BM25 is not in the top $N$ of DPR, we use the lowest score in DPR' top $N$, and vice versa. Ablation results on IHTest \citep{lin-etal-2019-kagnet} of CommonsenseQA and test set of OpenbookQA are shown in Table \ref{tab:ablation}.

When removing BM25, results on CommonsenseQA and OpenbookQA drop significantly up to 3.52\%. When removing DPR, the result on CommonsenseQA drops slightly while the result on OpenbookQA drops 3.20\%. These results show that BM25 and DPR are complementary, aligning with \citet{Ma2021ARS}. Similarly,  when further removing reranking module, model performance on CommonsenseQA drops 1.10\% and on OpenbookQA drops 4.00\% accuracy. These consistent results show the effectiveness of reranking  passages retrieved from hybrid retriever. In addition, the \underline{second best} model shown in our ablation study can still achieve strong performance and outperform best baseline GSC. These consistent results indicate that our model benefits from the combination of sparse and dense retrievers, and reranking module, even removing some of them can still have strong performance.

\section{Conclusion}

In this paper, we propose a simple but effective method for question answering over knowledge graphs with triplet retrieval. Extensive experiments on two datasets show that our method can consistently outperform state-of-the-art. Ablation study further shows that our model benefits from both reranking module and the combination of sparse and dense retrievers. We believe that our work can inspire future research for question answering over knowledge graphs. 

\section*{Limitations}
Our work is constrained into multi-choice question answering system and limited to common sense reasoning tasks, lacking more exploration in other reasoning tasks, e.g. arithmetic reasoning \citep{Cobbe2021TrainingVT,chen-etal-2021-finqa}, conversational reasoning \citep{Chen2022ConvFinQAET} and symbolic reasoning \citep{Wei2022ChainOT}. We plan to leave these directions as future work. 

\section*{Ethics Statement}

Our work utilizes pre-trained language model and external knowledge graph to build question answering systems. However, pre-trained language models can include biases \cite{shwartz-choi-2020-neural} and knowledge graph, e.g. ConceptNet, has been found to contain representational harms \cite{mehrabi-etal-2021-lawyers}, which can cause these question answering systems to inherit these potential biases and harms. Therefore, additional procedures, e.g. declining inappropriate inputs and filtering harmful outputs, must be taken before real-world deployment.

\bibliography{anthology,custom}
\bibliographystyle{acl_natbib}
\appendix
\section*{Appendix}
\section{Relation mapping}
\label{sec:relation_mapping}
We describe relation mapping with more details in Table \ref{tab:relation_mapping}, where left column shows the relation names and right column shows their corresponding templates we use to convert triplets into natural language sentences.
\begin{table}[h]
\small
\centering
\begin{tabular}{ll}
\toprule
relation name & relation template  \\
\midrule
    Antonym & is the antonym of \\ 
    AtLocation & is at location of \\
    CapableOf & is capable of \\
    Causes & causes \\
    CreatedBy & is created by \\
    IsA & is a kind of \\
    Desires & desires \\
    HasSubevent &has subevent \\
    PartOf &is part of \\
    HasContext &has context \\
    HasProperty &has property\\
    MadeOf &is made of\\
    NotCapableOf &is not capable of\\
    NotDesires &does not desire\\
    ReceivesAction &is\\
    RelatedTo &is related to\\
    UsedFor &is used for\\
    LocatedNear & is located near\\
    CausesDesire & causes the desire of\\
    MotivatedByGoal & is motivated by the goal of\\
    DistinctFrom & is distinct from\\
    HasFirstSubevent & has the first subevent\\
    HasLastSubevent & has the last subevent\\
    HasPrerequisite & has the prerequisite of\\
    Entails & entails\\
    MannerOf & a manner of\\
    InstanceOf & an instance of\\
    DefinedAs & is defined as\\
    HasA & has a\\
    SimilarTo & is similar to \\
    Synonym & is the synonym of \\
\bottomrule
\end{tabular}
\caption{Relation name mapping for ConceptNet. We adapt this relation mapping from \citet{yasunaga-etal-2021-qa}.}
\label{tab:relation_mapping}
\end{table}
\section{More implementation and training details} \label{sec:appendix}
We set $N=100$ for both CommonsenseQA and OpenbookQA in passage retrieval. 
We set $K=100$ and $K=20$ for CommonsenseQA and OpenbookQA, respectively, in the reranking step. Note that for CommonsenseQA, we applied additional rules after reranking stage to filter out passages. Specifically, we filter out passages that contain the "RelatedTo" relation or do not possess any token overlaps with the answer choices. We implement the method based on huggingface \texttt{transformers} \citep{wolf-etal-2020-transformers}, and train the models on NVIDIA A100-SXM4-40GB GPUs. For CommonsenseQA, we use the contextualized representation of first token from PLM as $\mathbf{h}$ while for OpenbookQA, we use the average contextualized representations of answer choice from PLM as $\mathbf{h}$. For both CommonsenseQA and OpenbookQA datasets, we use AdamW \citep{loshchilov2017adamw} with learning rate $10^{-5}$. We set the maximum training epoch as $15$ for CommonsenseQA and $10$ for OpenBookQA. We set batch size as $32$ and $16$ for CommonsenseQA and OpenBookQA, respectively. The maximum sequence length is set to be 512 for CommonsenseQA and OpenBookQA. We run experiments with three different random seeds $\{0, 1, 2\}$ and report mean results along with standard deviations.

\end{document}